\pdfoutput=1

\documentclass[11pt]{article}

\usepackage[final]{acl}

\usepackage{times}
\usepackage{latexsym}

\usepackage[T1]{fontenc}

\usepackage[utf8]{inputenc}

\usepackage{microtype}

\usepackage{inconsolata}

\usepackage{graphicx}
\usepackage{amssymb}
\usepackage{inconsolata}
\usepackage{amsmath,amssymb}
\usepackage{epsfig,graphicx,subfigure,caption}
\usepackage{algpseudocode}
\usepackage[normalem]{ulem}
\usepackage{amsmath}
\usepackage[linesnumbered,algoruled,boxed,noend]{algorithm2e}
\usepackage{xcolor}
\usepackage{color, colortbl}
\usepackage{multirow,booktabs, hhline}
\usepackage{amsmath, bm}
\usepackage[linesnumbered,algoruled,boxed,noend]{algorithm2e}
\usepackage{wrapfig}
\usepackage{verbatimbox}
\usepackage{kantlipsum}
\usepackage{fancyvrb}
\usepackage[htt]{hyphenat}
\usepackage{pifont}
\usepackage{bbding}
\newcommand{\cmark}{\ding{51}}%
\newcommand{\xmark}{\ding{55}}%

\newcommand{\our}{GraphBridge\xspace}
\definecolor{Gray}{gray}{0.92}
\newsavebox\CBox
\def\textBF#1{\sbox\CBox{#1}\resizebox{\wd\CBox}{\ht\CBox}{\textbf{#1}}}
\definecolor{dark2green}{rgb}{0.1, 0.65, 0.3}
\definecolor{dark2orange}{rgb}{0.9, 0.4, 0.}
\definecolor{dark2purple}{rgb}{0.4, 0.4, 0.8}
\newcommand{\first}[1]{\textBF{\textcolor{dark2green}{#1}}}
\newcommand{\second}[1]{\textBF{\textcolor{dark2orange}{#1}}}
\newcommand{\third}[1]{\textBF{\textcolor{dark2purple}{#1}}}

\title{Bridging Local Details and Global Context in Text-Attributed Graphs}

\author{
     \textbf{Yaoke Wang~\textsuperscript{1}\footnotemark[1]},
     \textbf{Yun Zhu\textsuperscript{1}\footnotemark[1]},
     \textbf{Wenqiao Zhang\textsuperscript{1}}\footnotemark[2],
     \textbf{Yueting Zhuang\textsuperscript{1}},\\
     \textbf{Yunfei Li\textsuperscript{2}},
     \textbf{Siliang Tang\textsuperscript{1}}
     \\
 \textsuperscript{1}Zhejiang University,   
 \textsuperscript{2}Ant Group
 \\
{\{wangyaoke, zhuyun\_dcd, wenqiaozhang, yzhuang\}@zju.edu.cn}\\
{qixiu.lyf@antgroup.com},
{siliang@zju.edu.cn}
}

\begin{document}
\maketitle
\renewcommand{\thefootnote}{\fnsymbol{footnote}}
\footnotetext[1]{These authors contributed equally to this work.}
\footnotetext[2]{Corresponding Author}

\begin{abstract}
Representation learning on text-attributed graphs (TAGs) is vital for real-world applications, as they combine semantic textual and contextual structural information.
Research in this field generally consist of two main perspectives: local-level encoding and global-level aggregating, respectively refer to textual node information unification (\emph{e.g.}, {using Language Models})  and structure-augmented modeling (\emph{e.g.}, using Graph Neural Networks).
Most existing works focus on combining different information levels but overlook the interconnections, \emph{i.e.}, the contextual textual information among nodes, which provides semantic insights to bridge local and global levels.
In this paper, we propose \our, a \textit{multi-granularity integration} framework that bridges local and global perspectives by leveraging contextual textual information, enhancing fine-grained understanding of TAGs. 
Besides, to tackle scalability and efficiency challenges, we introduce a graph-aware token reduction module. 
Extensive experiments across various models and datasets show that our method achieves state-of-the-art performance, while our graph-aware token reduction module significantly enhances efficiency and solves scalability issues. Codes are available at \href{https://github.com/wykk00/GraphBridge}{https://github.com/wykk00/GraphBridge}
\end{abstract}
\section{Introduction}\label{sec:intro}

\textbf{T}ext-\textbf{A}ttributed \textbf{G}raphs (TAGs), characterized by the association of nodes with text attributes~\cite{yang2021graphformers}, are prevalent in diverse real-world contexts.
In TAGs, nodes represent entities with textual information and edges capture relationships between entities,
\emph{e.g.}, social graphs where each user is accompanied by a textual description and paper
citation graphs where textual content is linked to each respective paper.
These relationships yield specialized and crucial insights that are fundamental for our understanding, thereby facilitating the resolution of subsequent tasks.
The utilization of TAGs empowers us to unlock new discoveries across various domains, including graph learning~\cite{zhang2024text} and information retrieval~\cite{seo2024unleashing}.

\begin{figure}
    \centering
    \includegraphics[width=1\linewidth]{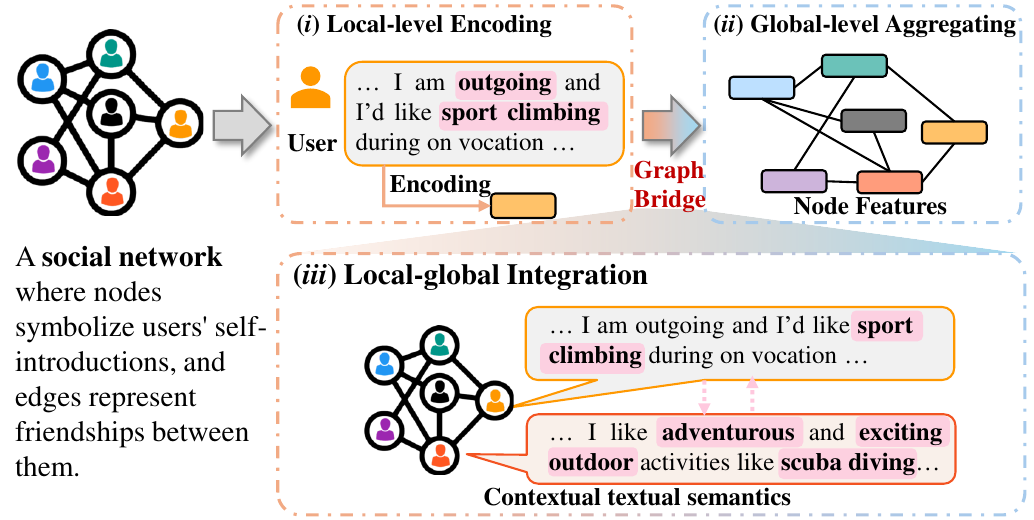}
   \caption{Illustration of the local-global integration in TAGs within a social network context. The words in pink emphasize the interconnection semantic relationship between them. (\textit{i}) The local-level encoding module processes individual nodes' textual information into unified vectors; (\textit{ii}) The global-level aggregating module enhances node features with structural information; (\textit{iii}) Our method bridges these two perspectives through incorporating contextual textual information.}
    \label{fig:introduction}
    \vspace{-0.5em}
\end{figure}

%
%
The nucleus of learning on TAGs lies in the effective integration of both the node attributes (textual semantics) and graph topology (structural connections) to facilitate the learning of node representations. 
Broadly, previous methods can be divided into two modules: (\textit{i}) encoding module (local-level) and (\textit{ii}) aggregating module (global-level). The encoding module transcribes the textual information (tokens) of each node into a unified vector employing static shallow embedding techniques such as Bag of Words~\cite{bow}, or language models (LMs) like BERT~\cite{bert}, serving as node attributes. The aggregating module enhances these features via structural information, procuring structure-augmented features through Graph Neural Networks (GNNs) like GCN~\cite{gcn}. These two modules can be integrated into cascading~\cite{duan2023simteg,giant}, joint~\cite{glem} or side structure~\cite{engine}, as illustrated in Figure~\ref{fig:introduction}.
Despite promising, the aforementioned local-global integration suffers from the discrete interconnection of encoding and aggregating modules, \emph{i.e.}, the contextual textual semantics among nodes are overlooked (Figure~\ref{fig:introduction}(\textit{iii})). In real-life scenarios, \emph{e.g.}, social networks, the closely connected individuals are more likely to share substantial semantically textual information in text, which serves as the common characteristics for constructing their relationships. 


Based on the aforementioned insight, one optimizable TAG learning solution is to leverage such contextual textual information that could effectively bridge local and global perspectives, thereby boosting fine-grained understanding of TAGs, like Figure~\ref{fig:introduction}(\emph{iii}). 
However, this method faces severe efficiency and scalability issues. Memory complexity increases with graph size, as neighborhood texts are also encoded. Using Large Language Models (LLMs) with densely connected nodes further exacerbates resource consumption, potentially impairing TAG's practicality.
In summary, these shortcomings necessitate a thorough reevaluation of TAG learning and its corresponding solutions.

In this work, we introduce a novel \textit{multi-granularity integration} framework for text-attributed graphs, named \our, which seamlessly bridges local and global perspectives by incorporating contextual textual information. This method enhances semantic analysis and provides deeper graph structure insights, significantly improving representation learning. 
Additionally, to address the efficiency and scalability issues mentioned above, we developed a graph-aware token reduction module. This module uses a learnable mechanism that considers both the graph structure and downstream task information to selectively retain the most crucial tokens, reducing information loss and allowing for the inclusion of more contextual text.
Extensive experiments show that our method achieves state-of-the-art performance across various domains compared to previous methods, while solving the efficiency and scalability issues. Key contributions of this work include:
\begin{itemize}
    \item We propose an innovative \textit{multi-granularity integration} framework named \our to integrate both local and global perspectives through leveraging contextual textual information, thereby enhancing the fine-grained understanding of TAGs.
    \item A graph-aware token reduction module is designed to ensure efficiency and scalability while minimizing information loss.
    \item  Extensive experiments conducted across various domains demonstrate that our proposed method achieves state-of-the-art performance compared to various baselines, demonstrating its effectiveness in bridging the gap between local and global information, while maintaining efficiency and scalability.
\end{itemize}

\section{Related Work}
\subsection{Representation Learning on TAGs}
Representation learning for text-attributed graphs has increasingly garnered attention in graph machine learning~\cite{yang2021graphformers}. Typically, previous methods in this field can be divided into two key components, as depicted in Figure~\ref{fig:introduction}: (\textit{i}) an encoding module at the local level, which employs word embedding methods such as Bag of Words~\cite{bow} or advanced LMs like BERT~\cite{bert} and RoBERTa~\cite{roberta} to generate token representations from nodes' textual data. These representations are integrated using methods like mean pooling to derive nodes' attributes; (\textit{ii}) an aggregating module at the global level, which utilizes GNNs~\cite{gcn,gat} or graph transformers~\cite{nodeformer} to augment nodes' attributes with structural information. 
The encoding module concentrates on extracting fine-grained semantic details from textual attributes individually, whereas the aggregating module emphasizes structural relationships between nodes, neglecting the intricate local textual information.

Recent advancements aim to effectively integrate these two modules. Integration strategies include joint frameworks~\cite{yang2021graphformers,glem} and side structures~\cite{engine}, as well as cascading approaches~\cite{duan2023simteg,tape}.
However, these integration strategies fail to explicitly capture the interconnection between the encoding and aggregating modules, \emph{i.e.}, the contextual textual semantics among nodes are frequently overlooked, potentially compromising the efficacy of the results.


\subsection{Token Reduction for LMs}
Sequence length has become a significant factor limiting the scalability of transformer models~\cite{vaswani2017attention}. Token reduction has gained considerable research interest because it can reduce computational costs by decreasing sequence length. The general idea of token reduction is to drop some tokens based on their importance~\cite{xu2023survey}. Specifically, DynSAN~\cite{zhuang2019token} applies a gate mechanism to measure the importance of tokens for selection, dropping less important tokens in higher layers to enhance efficiency. TR-BERT~\cite{trbert} introduces a dynamic reinforcement learning mechanism for making decisions of reducing tokens. LTP~\cite{kim2022learned} learns a threshold for each Transformer layer, dropping tokens with a saliency score below this threshold instead of adhering to a preset token reduction schedule. 

Although token reduction has proven successful in streamlining individual sentences, its application to interrelated texts within TAGs has yet to be fully explored. In this work, we propose a graph-aware token reduction module that leverages the graph structure along with the information from downstream tasks to perform token reduction.  
\section{Method}
In this section, we will introduce the notations used in this paper. Subsequently, we will present the proposed graph-aware token reduction method in Section~\ref{sec:reduction}. Finally, the multi-granularity integration framework will be discussed in Section~\ref{sec:framework}. 
\begin{figure*}
    \centering
    \includegraphics[width=1\linewidth]{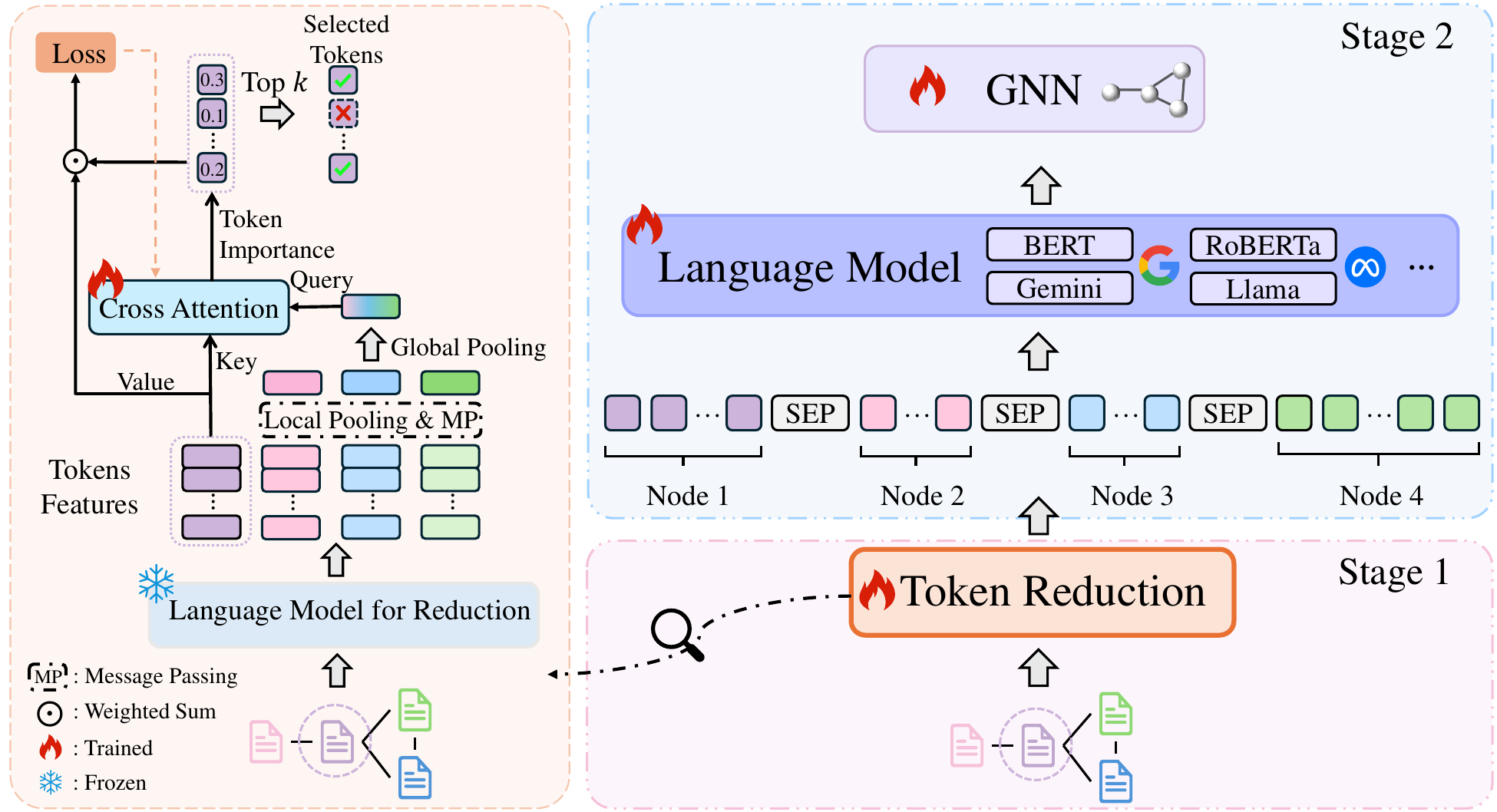} 
    \vspace{-1em}
    \caption{Overview of the \textbf{\our} framework. \textbf{Left}: The Graph-Aware Token Reduction module which selectively retains crucial tokens, enhancing efficiency and scalability. \textbf{Right}: A detailed pipeline illustrates the integration process, where selected tokens undergo a cascaded structure that bridges local and global perspectives, leveraging contextual textual information to effectively refine node representations.} 
    \label{fig:method}
    \vspace{-1em}
\end{figure*}

\subsection{Notations}
When dealing with node classification tasks of TAGs, we formally consider a text-attributed graph \(\mathcal{G}=\{\mathcal{V},\mathcal{T},\mathcal{A},\mathcal{Y}\}\),  where \(\mathcal{V}\) is a set of nodes, \(\mathcal{T}  \in \mathbb{R}^{|\mathcal{V}|\times k}\) denotes the textual features associated with each node \(i\in\mathcal{V}\), and \(k\) is the sequence length. 
 \(\mathcal{A}\in\{0,1\}^{|\mathcal{V}|\times|\mathcal{V}|}\) is the adjacency matrix where each entry \(\mathcal{A}_{i,j}\) indicates the link between nodes \(\ i,j\in\mathcal{V}\), and \(\mathcal{Y}\) represents labels for each node. 
Given a set of labeled nodes \(\mathcal{V}_{L} \subset \mathcal{V}\), our goal is to predict the remaining unlabeled nodes \(\mathcal{V}_{U} = \mathcal{V}\setminus \mathcal{V}_{L}\).

\subsection{Graph-Aware Token Reduction} \label{sec:reduction}
To bridge local and global perspectives for TAGs, it is essential to consider both the text of the current node and its neighboring nodes. This approach, however, leads to extremely long sequences that can result in prohibitive computational costs for LM processing. To mitigate this, we first implement a graph-aware token reduction module. Formally,  for an input graph $\mathcal{G}$, the text of each node \(\mathcal{T}_i\) is initially tokenized, resulting in \(s_i \in \mathbb{R}^k\), where \(k\) is the number of tokens. Our objective is to reduce the number of tokens to \(k'\) (where \(k'\ll k\)), focusing on retaining the most pivotal tokens while omitting the lesser ones. Specifically, we assess the importance of each token for node \(i\) based on its textual and structural information:
\begin{equation}
    P(\text{Score}_i \mid \mathcal{T}_i,\mathcal{T}_{\mathcal{N}_i},\mathcal{A}),
\end{equation}
where \(\text{Score}_i\in\mathbb{R}^{1 \times k}\) is the importance score for each token in node \(i\), and \(\mathcal{N}_i\) denotes the neighboring nodes of \(i\). The importance score is calculated by a trainable graph-enhanced attention module, as depicted in Figure~\ref{fig:method} (Stage 1).

For better evaluating the importance of each token, we first use Pre-trained LMs like BERT and RoBERTa to obtain fine-grained representations for each token. For node \(i\), these representations are represented as \({E}_{i} \in \mathbb{R}^{k \times d}\), consisting of vectors \([e_0, e_1, \ldots, e_k]\), where each \(e_j\) is a \(d\)-dimensional token representation from the PLM. We subsequently employ mean pooling $\mathcal{P}_{\text{mean}}$ on these token representations to extract the sentence-level textual feature, which serves as the node attribute:
\begin{equation}
    z_{i} = \mathcal{P}_{\text{mean}}(E_i)=\frac{1}{k}\sum_{j}^{k}e_{j}.
\end{equation}
Then, a parameter-free message-passing mechanism is employed to aggregate text features from neighboring nodes, excluding self-loops to avoid reinforcing a node's own information. This ensures the integration of contextual information from neighbors, enhancing node features with structural insights from graph. This process is described as:
\begin{equation}
    {z}_{i}^{(l)} = \frac{1}{| \mathcal{N}_i   |} \sum_{j \in \mathcal{N}_i} {z}_{j}^{(l-1)},
\end{equation}
where \(l\) means \(l\)-hop message passing, and \({z}_{i}^{0}\) is \(z_i\).

\paragraph{Graph-Enhanced Importance Score.} 
To measure the importance of each token, we define a graph-enhanced importance score calculated through a well designed cross-attention module, that quantifies the significance of each token utilizing both textual and structural information. Specifically, for each node \(i\), the query is derived from the message passing output of neighboring nodes, while the key and value come from the node's own textual token embedding. The importance score is calculated as follows: 
\begin{equation}
    \text{Score}_i=\sigma\left(\frac{\left( {z}_i^{(l)}W_{q}\right)\left({E}_iW_{k}\right)^{T}}{\sqrt{d}}\right),
\end{equation}
where \(W_{q},W_{k}\in\mathbb{R}^{d \times d^\prime}\) are the parameter matrices for the query and key, and \(\sigma\) denotes the softmax function. A top-k function is then used to select the \(k^\prime\) most crucial tokens.

\paragraph{Optimizing Importance Score.}
The supervisory signals derived from downstream tasks provide valuable guidance, allowing the attention module to select informative tokens more effectively. Specifically, during the training phase, we aggregate the token representations using \(\text{Score}_i\), which can be formulated as:
\begin{equation}
    s_i = \text{Score}_i{E}_i,
\end{equation}
where \(s_i\in\mathbb{R}^{1\times d}\) denotes the weighted summation of text features using attention score. Observe that \(E_i\) can be directly regarded as the value matrix, because we set the value parameter matrix as the identity matrix \(I\) here for efficiency. Finally, \(s_i\) is fed into a linear classifier $\mathcal{C}$ for prediction. The training loss $\mathcal{L}_{\text{down}}$ is computed using the cross-entropy loss $\operatorname{CE}(\cdot,\cdot)$ between the prediction and true label for the target node $i$:
\begin{equation}
    \mathcal{L}_{\text{down}}=\mathbb{E}_{i \in \mathcal{V}_{L}}\operatorname{CE}(\hat{y}_i\vert \mathcal{C}_i(s_i),y_i).
\end{equation}
Note that, only the attention module and the classifier are trained while keeping the PLM frozen for efficiency.

\begin{figure}
    \centering
    \includegraphics[width=1\linewidth]{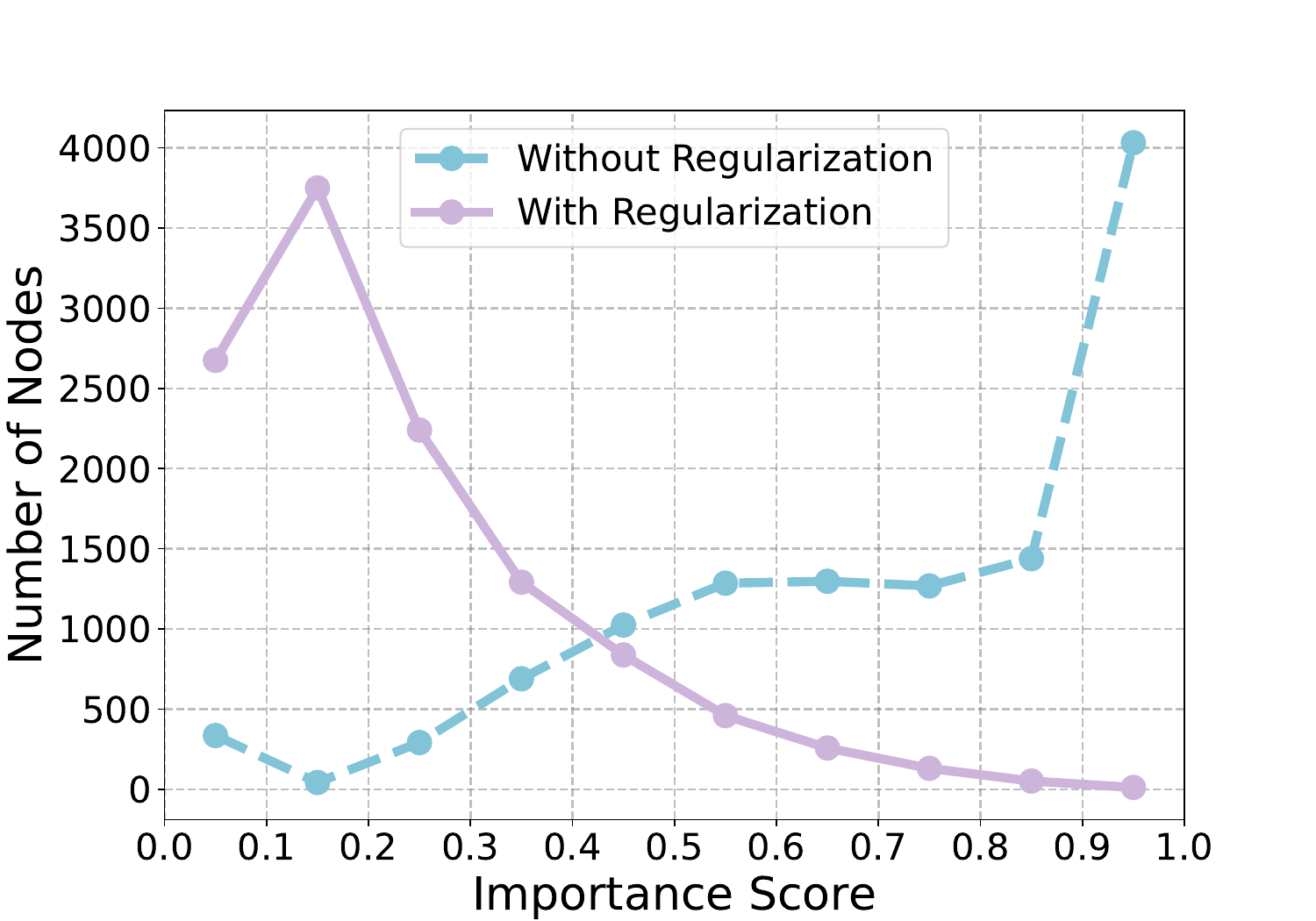}
    \vspace{-1em}
    \caption{Selecting the highest score token for each node in WikiCS dataset, with and without regularization. The x-axis means the highest importance score of token for each node, the y-axis indicates the number of nodes corresponding to each importance score.}
    \label{fig:reg_dist}
\end{figure}

\paragraph{Regularization.} 
Through our empirical study, we discovered that the importance score for the majority of nodes deteriorates when solely optimizing $\mathcal{L}_{\text{down}}$, as depicted in Figure~\ref{fig:reg_dist}. We hypothesize that this phenomenon is due to overfitting on the limited set of training nodes. Consequently, most nodes tend to converge on a single token with an excessively high importance score (\emph{e.g.}, 0.99), thereby hindering the selection of multiple informative tokens and inhibiting exploration. To mitigate this phenomenon, 
we introduce a regularization term. This term penalizes the network when certain tokens receive disproportionately high importance scores, achieved with a KL-divergence loss:
\begin{equation}
    \mathcal{L}_{\text{reg}} = \mathbb{E}_{i \in \mathcal{V}_{L}}D_{\text{KL}}(U\vert \vert\text{Score}_i),
\end{equation}
where \(U\) is an uniform distribution.

The overall training loss of the reduction model is as follows:
\begin{equation}\label{eqa:loss}
    \mathcal{L}_{\text{train}}=\mathcal{L}_{\text{down}}+\beta\mathcal{L}_{\text{reg}},
\end{equation}
where \(\beta\) is the regularization parameter for controlling the distribution of importance score. A larger \(\beta\) results in a more uniform score distribution, and vice visa.

\subsection{Multi-Granularity Integration} \label{sec:framework}
As discussed in Section~\ref{sec:intro}, the integration of local-global perspectives is essential for representation learning in TAGs~\cite{tagbenchmark}. However, previous methods~\cite{duan2023simteg,tape} often fail to adequately address the interconnection between these two perspectives. In this work, we propose an innovative multi-granularity integration framework that bridges local and global perspectives by considering contextual textual semantics among nodes, as illustrated in Figure~\ref{fig:method} (Stage 2). Intuitively, the contextual textual semantics among nodes can offer supplementary information for a given node. For example, within a citation network, the textual content of neighboring nodes might include key terms or concepts pertinent to the target node.

To incorporate contextual information, we consider both the text of each node and its neighbors. For each node \(i\), we concatenate its own text with the text of its neighboring nodes into a single sequence \(Q_i\):
\begin{equation}\label{eqa:seq}{Q}_i=(t_i^1,\cdots,t_i^{k_i^{\prime}},[\text{SEP}],t_{j_1}^1,\cdots,t_{j_n}^{k_n^{\prime}}),\end{equation}
where \(t^{k^\prime_i}_i\) represents the tokens in the node $i$, [SEP] is a separator token for separating different nodes, and \(\{j_1,\cdots,j_n\}\in \mathcal{N}_i\).
Note that concatenating the text of multiple nodes results in an excessively long sequence, which results in efficiency and scalability issues. Therefore, we utilize the token reduction module described in~Section~\ref{sec:reduction} to select the most crucial \(k^\prime\) tokens for forming the sequence.

After constructing the sequence \(Q_i\) for each target node $i$, we train the language model \(\operatorname{LM}(\cdot)\), which serves as the encoding module, on this sequence to obtain embeddings enriched with both textual and contextual information:
\begin{equation}
    \begin{aligned}
        \mathcal{L}_{\text{LM}}=\mathbb{E}_{i \in \mathcal{V}_{L}}\operatorname{CE}(\hat{y}_i\vert \mathcal{C}\left({\rm \operatorname{LM}}\left(Q_i\right)\right),y_i).
    \end{aligned}   
\end{equation}
It is noteworthy that sampling fewer neighboring nodes focuses the model more on the fine-grained semantic information from a local perspective, whereas sampling more neighboring nodes shifts the model's emphasis towards capturing the structural semantic information from a global perspective. 

After completing the training of the \(\text{LM}\), the model is used to produce node representations \(H\).  Subsequently, we train the aggregating module \(\operatorname{GNN}(\cdot)\) as follows:
\begin{equation}
    \begin{aligned}
      \!\!\! \!\! \mathcal{L}_{\text{GNN}}\!=\!\mathbb{E}_{i \in \mathcal{V}_{L}}\operatorname{CE}(\hat{y}_i|\mathcal{C}({\rm \operatorname{GNN}}(A,H)_i),y_i),\\
    \end{aligned}
\end{equation}
where the aggregating module \(\text{GNN}\) will generate node features that further reflect the structural semantics from a global perspective. 

Additionally, training the GNN and LM is fully decoupled, allowing for the use of any existing GNN and LM models. This cascading structure enables the independent optimization of each model, enhancing flexibility and facilitating integration with diverse architectures and applications.

\section{Experiments}
In this section, we first introduce the used datasets in Section~\ref{exp:datasets}. We then detail the baseline methods and experimental setup in Sections~\ref{exp:baselines} and~\ref{exp:setup}, respectively. Experiments are presented to evaluate our proposed method in Section~\ref{exp:main}. We further investigate the use of a causal large language model as the backbone in Section~\ref{exp:llm}. Finally, we provide an analysis of hyper-parameters, assess scalability and efficiency, and conduct an ablation study. Furthermore, additional experiments (\emph{e.g.}, link prediction) are in Appendix. 

\begin{table*}[ht]
    \centering
    \resizebox{0.8\linewidth}{!}{\begin{tabular}{lccccc}
        \toprule
        \textbf{Dataset} & \#\textbf{Nodes} & \#\textbf{Edges}   & \#\textbf{Avg.tokens} &\#\textbf{Avg.degrees}&\#\textbf{Classes}\\
        \midrule
        Cora& 2,708& 5,429&  194 &3.90&7\\
        WikiCS& 11,701& 215,863&  545 &36.70&10\\
        CiteSeer& 3,186& 4,277&  196 &1.34&6
\\
        ArXiv-2023& 46,198&  78,543&  253 &1.70&40\\
        Ele-Photo& 48,362&  500,928&  185 &18.07&12\\
 OGBN-Products (subset)& 54,025& 74,420& 163 &2.68&47\\
 OGBN-ArXiv& 169,343&  1,166,243& 231 &13.67& 40\\
 \bottomrule
    \end{tabular}
    }
    \caption{\textbf{Data statistics.} \#\textbf{Nodes}, \#\textbf{Edges}, \#\textbf{Classes} and \#\textbf{Avg.degrees} mean the number of nodes, edges,  classes and average degrees for each dataset, respectively. \#\textbf{Avg.tokens} represents the average number of tokens per node in each dataset when using the RoBERTa-base's tokenizer.}
    \vspace{-1em}
\end{table*}

\subsection{Datasets}\label{exp:datasets}
In this work, we adopt seven widely used textual graphs to evaluate our proposed \our: Cora~\cite{collective}, WikiCS~\cite{mernyei2020wiki}, CiteSeer~\cite{giles1998citeseer}, ArXiv-2023~\cite{tape},  Ele-Photo~\cite{tagbenchmark}, OGBN-Products~\cite{ogb} and OGBN-ArXiv~\cite{ogb}. The raw text of these datasets are collected by previous works~\cite{chen2023label,tagbenchmark,tape}. Details of these datasets can be found in Appendix~\ref{app:datasets}.

\begin{table*}[ht]
	\centering
        \resizebox{\textwidth}{!}{
		\begin{tabular}{l|ccccccc}
            \toprule
            Methods& Cora&WikiCS& CiteSeer& ArXiv-2023&Ele-Photo& OGBN-Products& OGBN-ArXiv\\
            \midrule
            MLP& 76.12 ± 1.51& 68.11 ± 0.76& 70.28 ± 1.13& 65.41 ± 0.16& 62.21 ± 0.17&	58.11 ± 0.23&	62.57 ± 0.11\\ 
            GCN& 88.12 ± 1.13&	76.82 ± 0.62&	71.98 ± 1.32&66.99 ± 0.19&	80.11 ± 0.09&	69.84 ± 0.52&70.78 ± 0.10\\
 SAGE& 87.60 ± 1.40& 76.65 ± 0.84& 72.44 ± 1.11& 68.76 ± 0.51& 79.79 ± 0.23& 70.64 ± 0.20&71.72 ± 0.21\\
 GAT& 85.13 ± 0.95& 77.04 ± 0.55& 72.73 ± 1.18& 67.61 ± 0.24& 80.38 ± 0.37& 69.70 ± 0.25&70.85 ± 0.17
\\
 NodeFormer& 88.48 ± 0.33& 75.47 ± 0.46& 75.74 ± 0.54& 67.44 ± 0.42& 77.30 ± 0.06&  67.26 ± 0.71&69.60 ± 0.08\\ 
            \midrule
            BERT& 79.70 ± 1.70&  78.13 ± 0.63& 71.92 ± 1.07& 77.15 ± 0.09 & 68.79 ± 0.11& 76.23 ± 0.19& 72.75 ± 0.09\\
            RoBERTa-base& 78.49 ± 1.36& 76.91 ± 0.69& 71.66 ± 1.18& 77.33 ± 0.16 & 69.12 ± 0.15& 76.01 ± 0.14& 72.51 ± 0.03\\
 RoBERTa-large& 79.79 ± 1.31& 77.79 ± 0.89& 72.26 ± 1.80& 77.70 ± 0.35& 71.22 ± 0.09& 76.29 ± 0.27&73.20 ± 0.13\\
             \midrule
              $\text{GLEM}_{(\text{base})}$& 87.61 ± 0.19&  78.11 
 ± 0.61& 77.51 ± 0.63& 79.18 ± 0.21& 81.47 ± 0.52&  76.15 ± 0.32& 74.46 ± 0.27\\

 $\text{TAPE}_{(\text{base})}$& 87.82 ± 0.91& $-$& $-$& 80.11 ± 0.20& $-$& 79.46 ± 0.11&74.66 
 ± 0.07\\

 $\text{SimTeG}_{(\text{base})}$& 86.85 ± 1.81& 79.77 ± 0.68
& 78.69 ± 1.12
& 79.31 ± 0.49& 81.61 ± 0.18& 76.46 ± 0.55&74.31 ± 0.14\\

 $\text{ENGINE}_{(\text{base})}$& 87.56 ± 1.48& 77.97 ± 0.94& 76.79 ± 1.38& 78.34 ± 0.15& 80.50 ± 0.33& 77.80 ± 1.20&73.59 ± 0.14\\
  \rowcolor{Gray}
    $\text{Ours}_{(\text{base})}$& \second{92.14 ± 1.03}& \second{80.59 ± 0.47}& \second{85.32 ± 1.39}& \second{84.07 ± 0.34}&\second{ 83.84 ± 0.07}&\second{ 79.80 ± 0.19}&74.89 ± 0.23\\
 
\midrule
 $\text{GLEM}_{(\text{large})}$& \third{89.11 ± 0.22}& 77.99 ± 0.72& 78.24 ± 0.31& 78.91 ± 0.40& 82.11 ± 0.66& 78.59 ± 0.27&74.98 ± 0.45\\
  $\text{TAPE}_{(\text{large})}$& 88.56 ± 0.88& $-$& $-$& 80.21 ± 
 0.31& $-$& \third{79.76 ± 0.23}&\second{75.29 ± 0.11}\\
  $\text{SimTeG}_{(\text{large})}$& 88.78 ± 1.05& 80.13 ± 0.76& 
\third{79.59 ± 1.56}& \third{80.51 ± 0.33}& 82.49 
 ± 0.17& 78.55 ± 0.66&\third{75.16 ± 0.21}\\
  $\text{ENGINE}_{(\text{large})}$& 88.49 ± 1.10& \third{80.21 ± 0.29}& 78.02 ± 0.87& 77.45 ± 0.46& \third{82.68 ± 0.09}&  78.83 ± 0.80&74.62 ± 0.30\\
  \rowcolor{Gray}
 $\text{Ours}_{(\text{large})}$& \first{92.73 ± 1.00}& \first{80.73 ± 0.41}& \first{86.81 ± 1.09}& \first{84.79 ± 0.29}& \first{84.18 ± 0.15}& \first{80.22 ± 0.47}&\first{75.90 ± 0.11}\\
            \bottomrule
 
        \end{tabular}
	}
 	\caption{\textbf{Experimental results of node classification}: We report the mean accuracy with a standard deviation of 5 runs with different random seeds. Highlighted are the top \first{first}, \second{second}, and \third{third} results. `base' and `large' refer to RoBERTa-base and RoBERTa-large as LM backbones, respectively. `$-$' indicates that datasets do not support for this method.}
  \label{tab:main}
  \vspace{-0.5em}
\end{table*}

\begin{table}[htpb]
	\centering
        \scalebox{0.75}{
		\begin{tabular}{l|ccc}
            \toprule
            Methods& Cora&WikiCS&Ele-Photo\\
            \toprule
 $\text{ LLaMA2}$&  82.80 ± 1.37& 80.82 ± 0.48&72.06 ± 0.10\\
    \midrule
 $\text{SimTeG}_{(\text{ LLaMA2})}$& 92.84 ± 0.13& 82.55 ± 0.51& 82.05 ± 0.17\\
 $\text{ENGINE}_{(\text{ LLaMA2})}$& 91.48 ± 0.32& 81.56 ± 0.97& 83.75 ± 0.08\\
    \midrule
\rowcolor{Gray}
 $\text{Ours}_{(\text{ LLaMA2})}$& \textbf{93.65 ± 0.44}& \textbf{84.18 ± 0.68}& \textbf{84.35 ± 0.12}\\
            \bottomrule
 
        \end{tabular}
	}
 	\caption{Experimental results when utilizing  LLaMA2-7B as the Large Language Model backbone. We employ LoRA with a rank of 4 to fine-tune the LLM and report the corresponding accuracy. We use \textbf{boldface} to denote the best performace.}
  \label{tab:LLM}
\end{table}

\subsection{Baselines} \label{exp:baselines}
To verify the effectiveness of our proposed method, we select several baseline models for comparison, categorized into three types:

\noindent\textbf{Traditional GNN-based methods}: primarily focus on the global level but utilize static shallow embeddings, which neglect fine-grained textual information, \emph{e.g.}, MLP, GCN~\cite{gcn}, SAGE~\cite{sage}, GAT~\cite{gat}, NodeFormer~\cite{nodeformer}.

\noindent\textbf{LM-based methods}: primarily focus on the local textual level and do not consider global structural information, ~\emph{e.g.}, BERT~\cite{bert}, RoBERTa~\cite{roberta}.

\noindent\textbf{Recent works designed for TAGs}: integrate both local and global levels, \emph{e.g.}, GLEM~\cite{glem}, TAPE~\cite{tape}, SimTeG~\cite{duan2023simteg}, ENGINE~\cite{engine}.

\subsection{Experimental Setup} \label{exp:setup}
For traditional GNN-based methods, we utilize the raw features of each dataset, which are derived using bag of words or one-hot vectors.
For LM-based methods, we fine-tune LMs with raw texts of each node on downstream tasks.
For recent TAGs methods, we select RoBERTa-base and RoBERTa-large as the LM backbones, and a two-layer SAGE with 64 hidden size as the GNN backbone.
Regarding to our method, we select the same LM and GNN backbones with recent TAGs methods for a fair comparison. Additionally, we utilize RoBERTa-base as the LM encoder for our token reduction module. For alternative LMs used in token reduction, please refer to Appendix~\ref{app:other}.

In our experiments, we fine-tune all parameters for base language models such as RoBERTa-base. For larger models like RoBERTa-large, we employ LoRA~\cite{hu2021lora} with a rank of 8 to ensure scalability and maintain consistency with the SimTeG approach~\cite{duan2023simteg}.

\subsection{Main Results}\label{exp:main}
From Table~\ref{tab:main}, we draw the following conclusions:

First, traditional GNN-based methods, which rely on global structural information using static shallow embeddings, underperform compared to current TAGs methods like GLEM that integrate both local and global levels information. For instance, on the ArXiv-2023 dataset, this integration results in a 12\% higher absolute performance over GNN methods such as GCN, highlighting the crucial role of local textual information in enhancing model efficacy.

Second, LM-based methods primarily focusing on local textual information fall short on TAGs, as evidenced by integration methods which utilize the same LM backbones surpassing them by about 10\% absolute performance on the Ele-Photo dataset, achieving over 80\% accuracy. This highlights the essential role of global structural information in creating more semantically and structurally aware node embeddings.

Last, our method surpasses existing local and global integration approaches for TAGs. Specifically, \our achieves an absolute improvement of over 6\% on the CiteSeer dataset and 4\% on the ArXiv-2023 dataset, outperforming the previous state-of-the-art method, SimTeG, across various LM backbones. This demonstrates the effectiveness of our approach in seamlessly integrating local and global perspectives by incorporating contextual textual information among nodes, thereby enhancing the fine-grained understanding of TAGs.

\subsection{Enhanced with Large Language Models}\label{exp:llm}

Our method, as demonstrated in Section~\ref{exp:main}, proves effective with small and medium discriminative LMs like RoBERTa-base and RoBERTa-large. Furthermore, we have expanded our method to incorporate causal Large Language Models (LLMs), which have shown significant capabilities across various natural language tasks~\cite{achiam2023gpt, zhang2024hyperllava, zhang2024revisiting}. Table~\ref{tab:LLM} presents the results obtained using LLaMA2-7B~\cite{touvron2023llama} as the LLM backbone. Our method outperforms both LM-based method~(\emph{i.e.}, LLaMA2) and integration methods~(\emph{i.e.}, SimTeG, ENGINE) that utilize LLM. This demonstrates the effectiveness of our method with the LLM backbone, highlighting the importance of bridging the local and global perspectives.

\subsection{Sensitive Analysis}

\paragraph{The number of walk steps.}
In the construction of the sequence \(Q\) as outlined in Equation~\ref{eqa:seq}, sampling neighboring nodes via a random walk with restart~\cite{rosa} is essential for effectively incorporating contextual textual information. The number of walk steps, a key hyper-parameter, dictates the extent of neighboring node inclusion and thus influences the breadth of contextual information captured.

\begin{figure}[htp]
    \centering
    \includegraphics[scale=0.53]{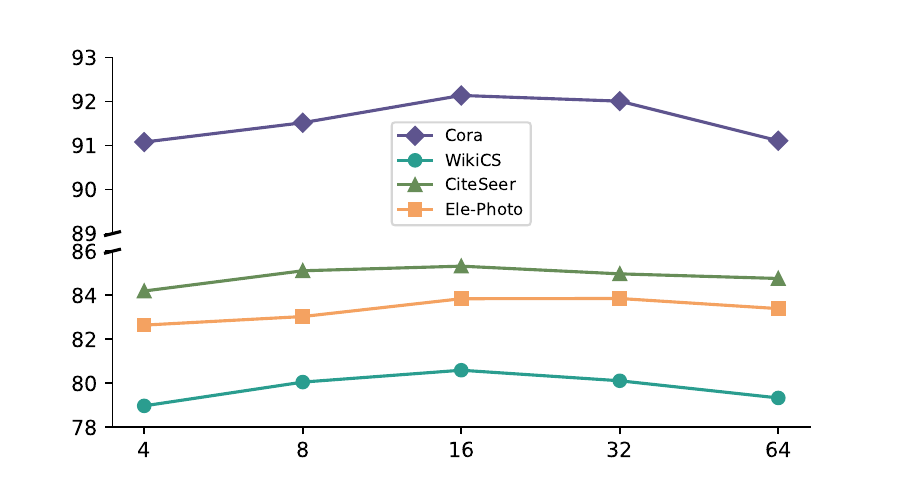} 
    \vspace{-1.2em}
    \caption{Sensitive analysis of the number of walk steps.}
    \label{fig:sensitive_walk}
\end{figure}

Empirically, we explore the impact of this number, choosing from \{4, 8, 16, 32, 64\}. The results in Figure~\ref{fig:sensitive_walk} indicate that a low number of walk steps like 4, leads to insufficient contextual information, while a high number like 64, may blur fine-grained local information and introduce noise, negatively impacting performance. To balance local and global information effectively, an intermediate number of steps, \emph{e.g.}, 16 or 32, is optimal. 

\begin{figure}[t]
    \centering
    \includegraphics[scale=0.5]{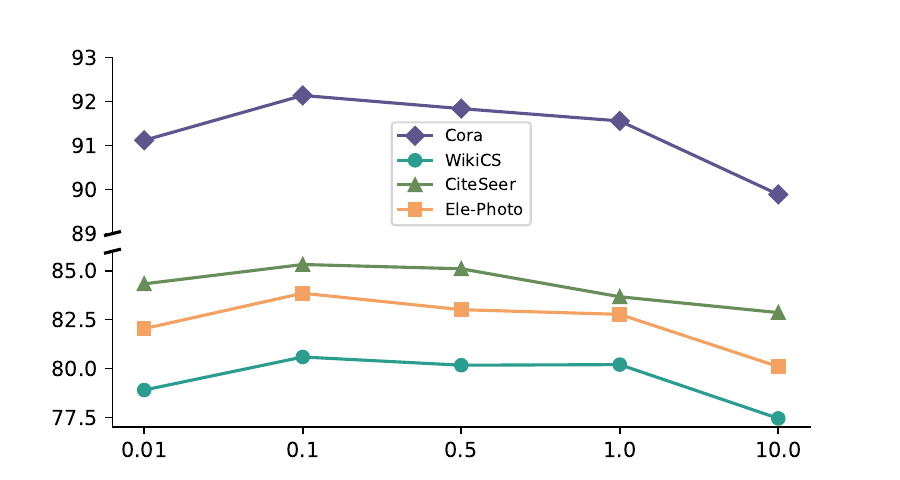} 
    \vspace{-1.2em}
    \caption{Sensitive analysis of regularization term \(\beta\).}
    \label{fig:sensitive_beta}
    \vspace{-0.5em}
\end{figure}

\paragraph{The regularization term \(\beta\).}
The regularization process penalizes the token reduction module when certain tokens receive extremely high importance scores. We analyze the regularization term~\(\beta\) as described in Equation~\ref{eqa:loss}, testing various value \{0.01, 0.1, 0.5, 1.0, 10.0\}. Based on Figure~\ref{fig:sensitive_beta}, optimal results are observed with a \(\beta\) value of 0.1. A small or large \(\beta\) value can lead to a steep or excessively smooth score distribution, respectively.

\subsection{Scalability and Efficiency Analysis}

\begin{table}[h]
    \centering
    \scalebox{0.85}{
    \begin{tabular}{c|c|c|c}
    \toprule
     Walk Steps & Reduction & Memory (GB) & Total Time \\ \midrule
     
     \multirow{2}{*}{8} & \xmark & 9.6& 39h 14m\\ 
      & \cmark &        3.2&      13h 55m\\ \midrule
      \multirow{2}{*}{16}& \xmark & 20.1&68h 42m\\ 
      & \cmark & 4.6& 18h 41m\\ \midrule
      \multirow{2}{*}{32}& \xmark & OOM& $-$\\ 
      & \cmark & 7.7& 31h 14m\\ 
      \midrule
      \multirow{2}{*}{64}& \xmark &OOM& $-$\\
      & \cmark &15.2&61h 51m\\
     \bottomrule
    \end{tabular}}
    \caption{The scalability and efficiency analysis of training on the OGBN-ArXiv dataset with and without token reduction. The batch size was set to 1 for LM tuning, with total training time reported using a 48-core Intel(R) Xeon(R) CPU @ 2.50GHz and 8 NVIDIA GeForce RTX 3090 GPUs. `OOM' refers to out of memory.}
    \label{tab:scale}
\end{table}

In this section, we assess the scalability and efficiency of our method by reducing the sequence length through token reduction module, detailed in Section~\ref{sec:reduction}. RoBERTa-base serves as our LM backbone, and for this experiment, we use the rotatory position embeddings~\cite{rope} to accommodate sequences of unlimited length. Results presented in Table~\ref{tab:scale}, demonstrate that without token reduction, method which considers neighboring textual information suffers from significant computational costs in terms of training time and GPU memory usage, and it may even run out of memory with a large number of walk steps (\emph{i.e.}, greater than 32).  With our token reduction module, we only retain the most~\(k'\) crucial tokens for each nodes, thereby enhancing efficiency and scalability.

\begin{figure}[t]
    \centering
    \includegraphics[scale=0.40]{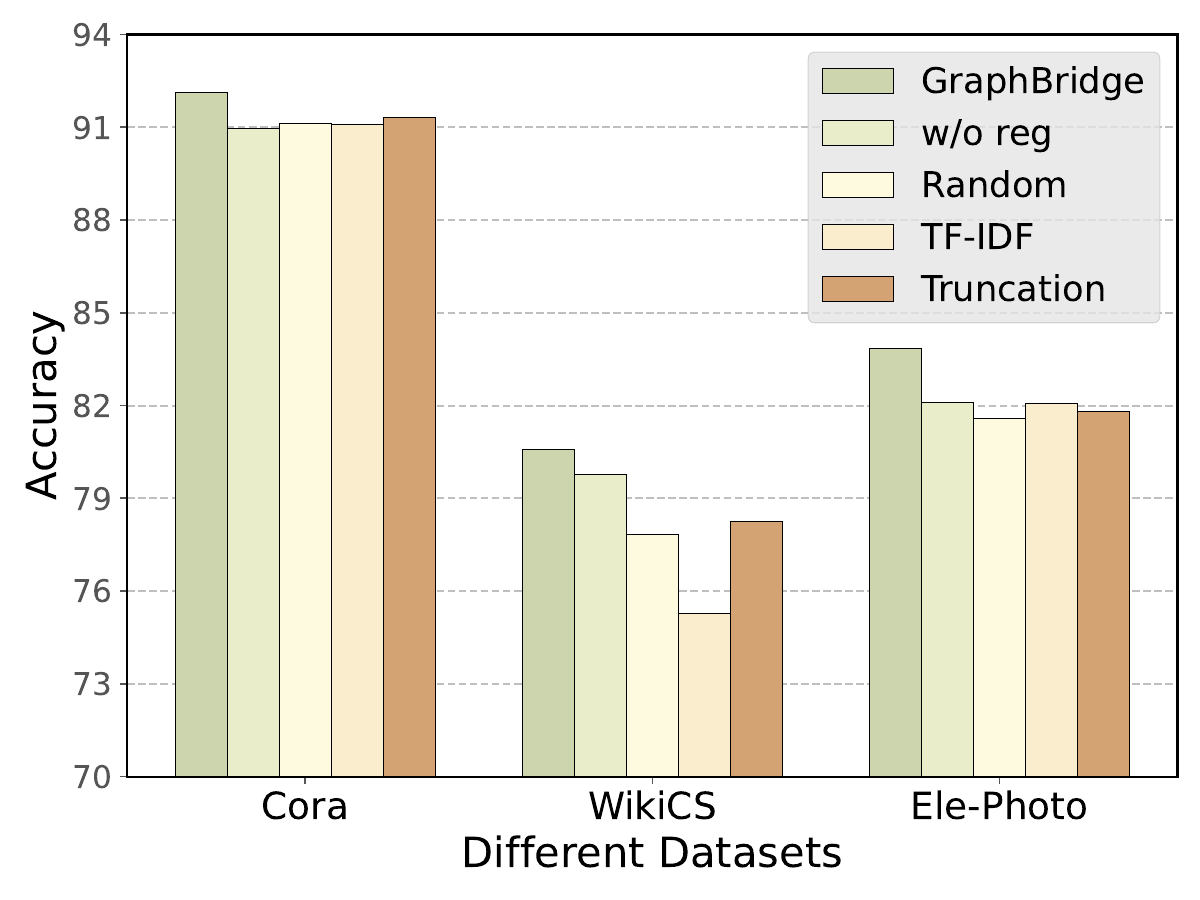} 
    \vspace{-2em}
    \caption{Experimental results of ablation study.}
    \label{fig:ablation}
    \vspace{-1em}
\end{figure}

\subsection{Ablation Study}
In this section, we evaluate the effectiveness of our token reduction module. Specifically, `w/o reg' refers to training the module without regularization. The methods `Random', `TF-IDF', and `Truncation' represent three different alternative token reduction strategies: random selection, selection based on TF-IDF scores, and selecting the initial tokens from texts, respectively. Figure~\ref{fig:ablation} shows that our token reduction module outperforms other reduction methods, highlighting its effectiveness in selecting crucial tokens. Additionally, regularization is essential as it helps prevent overfitting to specific tokens.
\section{Conclusion}
In this paper, we introduce \our, an innovative multi-granularity integration framework for text-attributed graphs. 
Our method emphasizes the importance of bridging local and global perspectives by incorporating contextual textual information, thereby enhancing the fine-grained understanding of TAGs. 
To tackle scalability and efficiency challenges associated with handling extensive textual data, we propose a graph-aware token reduction module.
Empirical studies confirm that \our surpasses existing state-of-the-art methods on various datasets.
\section{Limitations}
This work introduces a framework that seamlessly integrates both local and global perspectives by leveraging contextual textual information for TAGs. However, it primarily focuses on high-level discriminative tasks such as node classification and cannot be directly applied to generative tasks like graph description. Leveraging this framework to construct a graph foundation model presents a challenging yet valuable area for exploration.

\section{Acknowledgements}
This  work  has  been  supported  in  part  by  the  Key Research and Development Projects in Zhejiang Province (No. 2024C01106), the NSFC (No. 62272411), the National Key Research and Development Project of China (2018AAA0101900), and Ant Group.

\bibliography{custom, anthology}

\appendix
\clearpage

\section{Datasets}\label{app:datasets}
We evaluated our method using seven widely recognized text-attributed graph datasets. The details of these datasets are as follows:

\textbf{Cora}~\cite{collective} dataset contains 2,708 scientific publications classified into seven classes: case-based, genetic algorithms, neural networks, probabilistic methods, reinforcement learning, rule learning, and theory. Each paper
in the citation network cites or is cited by at least one other paper, resulting in a total of 5,429 edges.

\textbf{WikiCS}~\cite{mernyei2020wiki} dataset is a Wikipedia-based dataset designed for benchmarking Graph Neural Networks, consisting of 10 computer science branches as classes with high connectivity. Node features are derived from the corresponding article texts\footnote{We obtain the raw texts of each node from https://github.com/pmernyei/wiki-cs-dataset.}. 

\textbf{CiteSeer}~\cite{giles1998citeseer} dataset comprises 3,186 scientific publications categorized into six areas: Agents, Machine Learning, Information Retrieval, Database, Human Computer Interaction, and Artificial Intelligence, with the task of classifying each paper based on its title and abstract.

\textbf{ArXiv-2023} dataset, introduced in TAPE~\cite{tape}, is a directed graph representing the citation network of computer science arXiv papers published in 2023 or later. Similar to OGBN-ArXiv, it features nodes representing arXiv papers and directed edges for citations. The goal is to classify each paper into one of 40 subject areas such as cs.AI, cs.LG, and cs.OS, with classifications provided by the authors and arXiv moderators.

\textbf{Ele-Photo}~\cite{tagbenchmark} dataset, derived from the AmazonElectronics dataset~\cite{ni2019justifying}, consists of nodes representing electronics products, with edges indicating frequent co-purchases or co-views. Each node is labeled according to a three-level classification of electronics products. The text attribute for each node is the user review with the most votes, or a randomly selected review if no highly-voted reviews are available. The task is to classify these products into 12 categories.

\textbf{OGBN-Products}~\cite{ogb} dataset, comprising 2 million nodes and 61 million edges, is reduced using a node sampling strategy from TAPE~\cite{tape} to create the OGBN-Products (subset) with 54k nodes and 74k edges. Each node represents an Amazon product, with edges denoting co-purchases. The classification task involves categorizing products into one of 47 top-level categories.

\textbf{OGBN-ArXiv} dataset is a directed graph depicting the citation network among computer science arXiv papers indexed by MAG~\cite{wang2020microsoft}. Each node represents an arXiv paper with directed edges indicating citations. The goal is to classify papers into one of 40 subject areas like cs.AI, cs.LG, and cs.OS, with labels manually assigned by the authors and arXiv moderators. 

\begin{table*}[htpb]
	\centering
        \scalebox{0.9}{
		\begin{tabular}{l|cccc}
            \toprule
            & Cora&WiKiCS&CiteSeer&Ele-Photo\\
    \midrule
\rowcolor{Gray}
 $\text{Ours}_{(\text{ base})}$& 92.03 ± 0.94&  80.13 ± 0.31&84.52 ± 1.17& 83.14 ± 0.10\\
            \bottomrule
 \rowcolor{Gray}
 $\text{Ours}_{(\text{ large})}$& 91.96 ± 0.77&  80.55 ± 0.24&85.91 ± 0.99&84.34 ± 0.14\\
 
        \end{tabular}
	}
 	\caption{Results using BERT as an alternative language model backbone for token reduction.}
  \label{tab:other_lm}
\end{table*}

\section{Baselines}\label{app:baselines}
The details of the baseline methods we compared \our to are as follows:
\begin{itemize}
    \item Traditional GNNs: In this work, we adopted three simple yet widely used GNN models: GCN~\cite{gcn}, SAGE~\cite{sage}, and GAT~\cite{gat}. Additionally, We include a graph transformer as the GNNs-based methods baseline,~\emph{i.e.}, NodeFormer~\cite{nodeformer}.
    \item Fine-tuned Language Models: We adopt three commonly used pre-trained language models in our study: BERT~\cite{bert}, two versions of RoBERTa~\cite{roberta}, specifically RoBERTa-base and RoBERTa-large.
    \item Previous methods for TAGs: \textbf{GLEM}~\cite{glem} is an effective framework that fuses language models and GNNs in the training phase through a variational EM framework. We use the official source code\footnote{https://github.com/AndyJZhao/GLEM} to reproduce its results. \textbf{TAPE}~\cite{tape} utilizes Large Language Models like ChatGPT~\cite{achiam2023gpt} to generate pseudo labels and explanations for textual nodes, which are then used to fine-tune Pre-trained Language Models alongside the original texts. We reproduced its results using the official source code\footnote{https://github.com/XiaoxinHe/TAPE }. \textbf{SimTeG}~\cite{duan2023simteg} employs a cascading structure specifically designed for textual graphs, utilizing a two-stage training paradigm. Initially, it fine-tunes language models and subsequently trains GNNs. We conducted experiments using the official source code\footnote{https://github.com/vermouthdky/SimTeG}. \textbf{ENGINE}~\cite{engine} is an efficient fine-tuning and inference framework for text-attributed graphs. It co-trains large language models and GNNs using a ladder-side approach, optimizing both memory and time efficiency. For inference, ENGINE utilizes an early exit strategy to further accelerate. We reproduce its results using the official source code\footnote{https://github.com/ZhuYun97/ENGINE}.
    
\end{itemize}

\section{Implementation Details}
In this section, we give the implementations details about our method~\our. 

\paragraph{Training of Graph-Aware Token Reduction Module.} We train only the cross-attention module and the classifier, keeping the encoding language models frozen~\cite{zhang2019frame, zhang2022boostmis}. Each dataset undergoes 100 training epochs, with an early stopping patience of 10 epochs. The learning rate is explored within \{1e-3, 5e-4, 1e-4\}, and the regularization term~\(\beta\) is set to 0.1.

\paragraph{Training of Language Models.} Initially, we construct the sequence~\(Q\) by sampling adjacent nodes using a random walk with restart sampler~\(\Gamma\)~\cite{rosa,engine}. The number of walk steps for sampling varies among~\{8, 16, 32\}. Training epochs for the language models are adapted according to the dataset sizes: \{4, 6, 8\} for small datasets (\emph{e.g.}, Cora, WikiCS, CiteSeer), \{4, 6\} for medium datasets (\emph{e.g.}, ArXiv-2023, Ele-Photo, OGBN-Products), and \{4\} for the large-scale dataset (OGBN-ArXiv). We employ AdamW optimizers. The learning rate is explored within \{1e-4, 5e-5, 1e-5\} for full parameter fine-tuning of RoBERTa-base\footnote{https://huggingface.co/FacebookAI/roberta-base}, and \{1e-3, 5e-4, 1e-4\} for tuning RoBERTa-large\footnote{https://huggingface.co/FacebookAI/roberta-large} using LoRA~\cite{hu2021lora} with a rank of 8. For the large language model LLaMA2-7B\footnote{https://huggingface.co/meta-llama/Llama-2-7b}, as outlined in Table~\ref{tab:LLM}, we use LoRA with a rank of 4 and a learning rate within \{5e-4, 1e-4, 5e-5\}.

\paragraph{Training of Graph Neural Networks.} We train the GNNs models~(\emph{i.e.}, SAGE) subsequent to acquiring node representations from the language models. Specifically, The number of training epochs is designated within the range of \{100, 200, 500\}, complemented by an early stopping mechanism set at 20 epochs for each dataset. We utilize the Adam optimizers, and the learning rate is chosen from the set~\{1e-2, 5e-3, 1e-3\}.

\section{Alternative LMs as Backbones for Token Reduction}\label{app:other}
Our graph-aware token reduction module is compatible with any language model as a backbone. In this section, we demonstrate the effectiveness of our token reduction module using BERT\footnote{https://huggingface.co/google-bert/bert-base-uncased}~\cite{bert} as an alternative backbone. Table~\ref{tab:other_lm} presents the results when employing BERT for token reduction.

\section{Link Prediction}
\begin{table}[htpb]
	\centering
        \scalebox{0.75}{
		\begin{tabular}{l|ccc}
            \toprule
            Methods& Cora&CiteSeer	&ArXiv-2023\\
            \toprule
 $\text{SAGE}$&  	86.20 ± 0.96		& 79.69 ± 1.11&94.90 ± 0.24\\
    \midrule
 $\text{RoBERTa-base}$& 83.68 ± 0.61	& 85.10 ± 1.23& 92.59 ± 0.55\\
    \midrule
 $\text{SimTeG}$& 87.24 ± 0.47		& 86.33 ± 1.05& 97.01 ± 0.64\\
\rowcolor{Gray}
 $\text{Ours}_{(\text{base})}$& 	\textbf{88.87 ± 0.81}		& \textbf{88.01 ± 1.20}& \textbf{97.33 ± 0.48}\\
            \bottomrule
 
        \end{tabular}
	}
 	\caption{Experimental results of link prediction. The mean AUC with standard deviation across 5 runs is reported.}
  \label{tab:link}
\end{table}

Our method is not limited to node classification tasks, it can also be applied to other representation learning tasks of TAGs (\emph{e.g.}, link prediction). In this section, we conduct experiments on link prediction task. Specifically, we split the existing edges into train:val:test=0.6:0.2:0.2 for all datasets. The AUC score is served as the metric. According to Table~\ref{tab:link}, \our still outperforms other baselines by a significant margin. This highlights the effectiveness of its representation learning on TAGs and demonstrates the potential of GraphBridge for application to other-level tasks.


\end{document}